\documentclass{article}

\usepackage{microtype}
\usepackage{graphicx}
\usepackage{subfigure}
\usepackage{multirow}       
\usepackage{booktabs} 

\usepackage{hyperref}

\usepackage[accepted]{icml2023}

\usepackage{amsmath}
\usepackage{amssymb}
\usepackage{mathtools}
\usepackage{amsthm}

\usepackage[capitalize,noabbrev]{cleveref}

\theoremstyle{plain}

\theoremstyle{definition}

\theoremstyle{remark}

\usepackage[disable,textsize=tiny]{todonotes}

\icmltitlerunning{Physics-Informed Localized Learning for Advection-Diffusion-Reaction Systems}

\begin{document}

\twocolumn[
\icmltitle{Physics-informed Localized Learning for Advection-Diffusion-Reaction Systems}



\icmlsetsymbol{equal}{*}

\begin{icmlauthorlist}
\icmlauthor{Surya T. Sathujoda}{cam}
\icmlauthor{Soham M. Sheth}{slb}
\end{icmlauthorlist}

\icmlaffiliation{cam}{Department of Applied Mathematics and Theoretical Physics, University of Cambridge, Cambridge, UK}
\icmlaffiliation{slb}{SLB, Abingdon, UK}

\icmlcorrespondingauthor{Surya T. Sathujoda}{sts40@cam.ac.uk}

\icmlkeywords{Deep Learning, Embed to Control, Reservoir surrogate modelling, Localised Learning, Physics-informed Machine Learning}

\vskip 0.3in
]



\printAffiliationsAndNotice{} 

\begin{abstract}
The global push to advance Carbon Capture and Sequestration initiatives and green energy solutions, such as geothermal, have thrust new demands upon the current state-of-the-art subsurface fluid simulators. The requirement to be able to simulate a large order of reservoir states simultaneously, in a short period of time, has opened the door of opportunity for the application of machine learning techniques for surrogate modelling. We propose a novel physics-informed and boundary condition-aware Localized Learning method which extends the Embed-to-Control (E2C) and Embed-to-Control and Observe (E2CO) models to learn local representations of global state variables in an Advection-Diffusion Reaction system. Trained on reservoir simulation data, we show that our model is able to predict future states of the system, for a given set of controls, to a great deal of accuracy with only a fraction of the available information. It hence reduces training times significantly compared to the original E2C and E2CO models, lending to its benefit in application to optimal control problems.
\end{abstract}

\section{Introduction}

Subsurface fluid simulators play an important role in modelling various modern geological processes. They enable the efficient and sustainable utilization of geothermal energy by modelling heat and mass transfer processes; they play a vital role in modern Carbon Capture and Sequestration (CCS) initiatives by assessing optimal locations to inject $CO_2$ into geological formations, and they facilitate the efficient extraction of natural resources to reduce the risks of exploration. Developments in these areas have produced novel challenges to the production workflow of subsurface fluid modelling. The task of optimizing the location and volume of $CO_2$ injection into a porous medium, for instance, requires vast amounts of simulation data and compute power for each individual control state and time step of the system to determine the maximal storage case. This is especially problematic for applications in which we require real-time decision making. This has hence given birth to a new class of subsurface surrogate modelling techniques using machine learning.

The governing equation of subsurface flow, derived from the law of mass-conservation and Darcy's law, relates the fluid flow and flow potential gradients of a multi-phase porous-medium as given below

\begin{center}
\begin{math}
\nabla \cdot \biggl[ \alpha \textbf{k} \, \frac{k_{r,m}(S_m)}{\mu_m B_m}(\nabla p_m - \gamma_m \nabla z) \biggr] - \beta \frac{\partial}{\partial t} \biggl( \frac{\phi S_m}{B_m} \biggr) + \sum_w \frac{q_{sc,m}^w}{V_b} = 0 
\end{math}
\end{center}

where $\textbf{k}$ denotes the permeability tensor, $k_r$ the relative permeability, $\mu$ the viscosity, $B$ the formation volume factor, $p$ the pressure, $S$ the saturation, $\phi$ the porosity, $t$ the time, $\gamma$ the specific weight, $q$ the source/sink terms, $z$ the depth and $V_b$ the bulk volume. The subscript $sc$ represents standard conditions and $\alpha$ and $\beta$ are unit field constants. More generally, this equation is an instance of an Advection-Diffusion-Reaction (ADR) partial differential equation (PDE) which describes the transport and transformation of scalar quantities (such as pressure and saturation in this case) in a medium due to advection, diffusion and reaction, as the name suggests. The ADR equation arises in many other areas of science and engineering such as environmental engineering, where it can be applied to help understand the spread of contaminants in groundwater, biomedical engineering, where it can be applied to model the transport of drugs in biological tissues, and in chemical engineering, where it can be used to help optimize reactor design and operation. Although our proposed model focuses on subsurface fluid dynamics, it easily generalises to all the above areas with the utilisation of the appropriate domain specific knowledge. 

The above equation of subsurface flow describes a highly non-linear system that even state-of-the-art numerical solvers using Newton's method require significant compute power to converge to a solution. For this reason, various reduced-order-models (ROMs) have been developed which approximate the evolution of state variables in time to be linear. The most common such ROM is the Proper Orthogonal Decomposition Trajectory Piece-wise Linearization Reduced Order Model (POD-TPWL) \cite{TPWL1, TPWL2, TPWL3, TPWL4, TPWL5, TPWL6, TPWL7} which reduces the size of the system-state variables via the POD step before approximating their evolution linearly for each time step of the control. Although this method has played an important role in speeding up calculations, in recent years machine learning methods have been developed which, with a one time overhead training cost, promise an even faster inference of the next system state.

Early on, some general machine learning models were first applied to this problem by treating the initial field values at each point as the input to the model and the numerical solution to the PDE as the output for time-independent cases and the next step field values as the output for time dependent cases. These were purely data-driven approaches which were agnostic to the form or the physics of the underlying PDEs governing the system. Some of these early methods, which now serve as a benchmark for which to compare more advanced methods to, are: \textbf{NN}: a simple point-wise feedforward neural network. \textbf{RBM}: the classical Reduced Basis Method (using POD basis). \textbf{FCN}: a state-of-the-art neural network architecture based on Fully Convolution Networks \cite{ZHU2018415}. \textbf{PCANN}: an operator method using PCA as an auto-encoder on both the input and output data and interpolating the latent spaces with a neural network \cite{bhattacharya2021model}. For time dependent tasks, popular models in other time dependent regression tasks were applied such as: \textbf{ResNet}: a residual learning framework to ease the training of networks that are substantially deeper than those used previously \cite{he2015deep}. \textbf{U-Net}: A popular choice for image-to-image regression tasks consisting of four blocks with 2-d convolutions and deconvolutions \cite{ronneberger2015unet}.

As developments continued, one of the main directions of success to model non-linear control dynamics came from the Embed-to-Control (\textbf{E2C}) model \cite{E2C}. The E2C model consists of a variational auto-encoder and a transitional block which learns to generate image trajectories from a latent space in which the dynamics are constrained to be locally linear. This work was subsequently modified by replacing the variational auto-encoder with an ordinary encoder-decoder structure and then applied to the problem of 2D reservoir surrogate modelling by \cite{E2C2}. The results produced by this model were extremely accurate compared to previous methods and formed the bed rock for the expansion of the model to predict well outputs also for resource extraction in \cite{E2CO}. The thus proposed Embed to Control and Observe (\textbf{E2CO}) model added an additional parameterized network flow in the transitional block to predict well outputs such as flow rates at source/sink points. Both the E2C and the E2CO models were naturally expanded to 3D grids by \cite{E2C3} by remodelling the architecture to involve 3D convolution blocks and transposes in the encoder and decoder respectively to more reflect real-world systems in the area. The main draw back of these current models however is that, although they produce promising predictions for future time steps, the training times grow rapidly as grid sizes increase. Training time for a 3D case of $5 \times 10^5$ cells, which is on the lower end of a real-world case, takes upwards of 47 hours on a NVIDIA Tesla V100 GPU. Various new architectures have been proposed to tackle this problem, such as \textbf{Neural Operators} \cite{li2020neural, li2020multipole, li2021fourier, li2022physicsinformed, kovachki2023neural} and \textbf{DeepONet} \cite{DeepO2021}, but in their current state, they do not rival the accuracy of convolution-based methods.

We hence propose Localized Learning for Embed to Control (\textbf{LL-E2C}) and Localized Learning for Embed to Control and Observe (\textbf{LL-E2CO}), a method that learns on a random set of sub-grids of the original full grid of simulation, using physics-informed losses, and reconstructs the next state using the locally learned model. The intuition behind this being that physics at a local level is constant everywhere when boundary conditions are accounted for. We show that we achieve similar levels of accuracy to E2C and E2CO with our model while drastically reducing the training times.

\section{Preliminaries}
\begin{figure}
  \centering
  \includegraphics[scale=0.4, trim={15cm 0 15cm 0cm}]{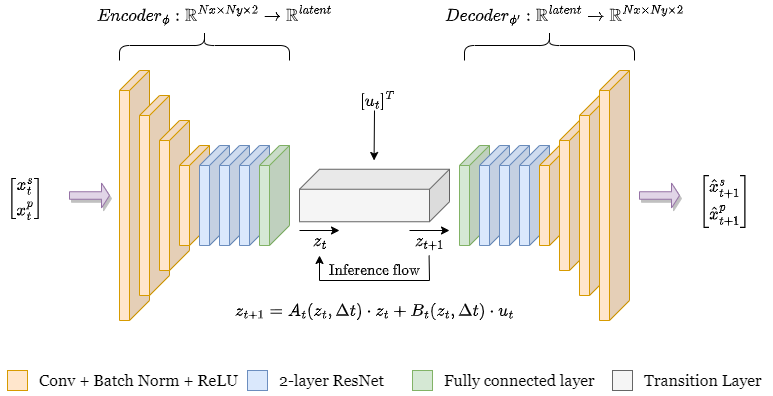}
  \caption{2D architecture for Embed-to-control applied to the reservoir simulation problem where $x^s_t$ and $x^p_t$ represent the saturation and pressure values respectively and $u_t$ the well controls.}
  \label{fig:method_2D}
\end{figure}

The task that we apply our model to is that of a reservoir simulation. The reservoir is discretized into a regular grid and different geological structures are modelled by spatially varying but temporally static permeabilities across the grid. At each point in the grid we have a certain pressure and saturation value and at various points in the reservoir we have sources/sinks in the form of injection/production wells. We can control these wells to affect the source an sink terms in the governing equation which in turn aids an important application of reservoir simulation, that is the deduction of optimal control and well placement for a given subsurface which maximises well production and storage rates. 

\subsection{Problem Setting}
Let a fluid dynamical system be discretized on a regular grid with dimensions $n = N_x \times N_y \times N_z$. The state of the system at time $t$ is represented by $\textbf{x}_t \in \mathcal{G}$, where $\mathcal{G} \subset \mathbb{R}^{n \times d_x}$. Given an applied control $\textbf{u}_t \in \mathcal{U}$, where $\mathcal{U} \subset \mathbb{R}^{d_u}$, we aim to model the arbitrary, smooth, system dynamics function $f$ in the time evolution equation $\textbf{x}_{t+1} = f(\textbf{x}_t, \textbf{u}_t)$ and the function $g$ determining the well outputs $\textbf{y}_{t+1} \in \mathcal{Y} $ where $\mathcal{Y} \subset \mathbb{R}^{d_u}$, given by $\textbf{y}_{t+1} = \bf{g}(x_t, u_t)$.

\subsection{Data}
Training and testing data is generated by a proprietary high-fidelity reservoir simulator (HFS). We consider two separate cases to evaluate our model, a 2D ($60 \times 60$) case and a larger 3D ($60 \times 220 \times 40$) case. In the 2D case, each train/test sample consists of pressure $\textbf{p}_t \in \mathbb{R}^n$, saturation $\textbf{S}_t \in \mathbb{R}^n$ and control $\textbf{u}_t$ which corresponds to the source/sink terms for $T=24$ times steps. We also have well output values for the next time step $\textbf{y}_{t+1}$ in the case of E2CO. The input to the model is constructed such that $\textbf{x}_t = [\textbf{p}_t, \textbf{S}_t]$. Permeability and source/sink location information, which is static with time, are available where required. The same data is simulated for the 3D case but with $T=20$. We generate 400 such samples and utilise a 3:1 split for training and testing.

\section{Methodology}
\begin{figure}
  \centering
  \includegraphics[scale=0.3, trim={0cm 0 0cm 0}]{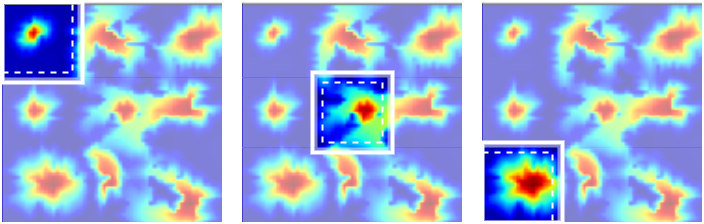}
  \caption{Three different cases of inducing boundary condition awareness in the model. The solid line indicates the sector which is used in training and inference and the cells inside the dotted lines indicate the part of the sector which are `stitched' together to recreate the full next state. \textbf{Left.} The stitched sub-sector shares hard boundaries on the top and left with the actual grid, teaching the model that flow cannot continue in that direction. \textbf{Middle.} The stitched sub-sector shares no hard boundaries allowing the model to learn that fluid can flow out of the sub-sector in all directions. \textbf{Right.} The sub-sector shares the same hard boundaries on the left and bottom as the grid, imposing Dirichlet boundary conditions, and Von Neumann on the right and top.}
  \label{fig:bound_method}
\end{figure}

\subsection{E2C Reduced Order Model}

\begin{figure*}[!htb]
  \centering
  \includegraphics[scale=0.5, trim={2cm 1cm 0cm 0}, width=\textwidth]{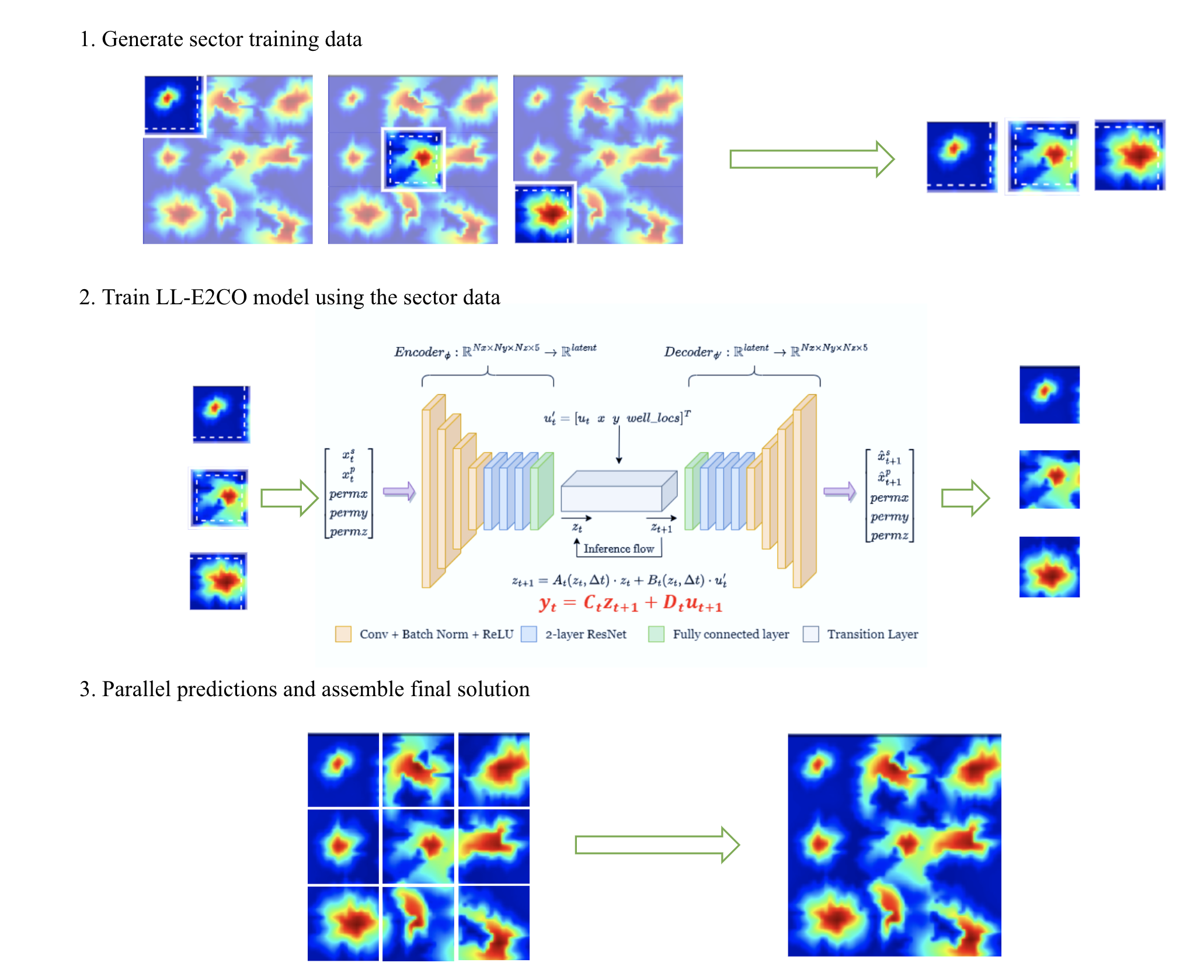}
  \caption{\textbf{Top.} Boundary condition-aware mechanism to train and infer on the sub-grid within the solid white line but stitch up using the sub-grid within the dotted white line. \textbf{Middle.} Model architecture of the 3D LL-E2C with the equation in red showing the additional flow for LL-E2CO model. \textbf{Bottom.} Representative next state prediction using individual stitching of sectors to reconstruct the full image.}
  \label{fig:method}
\end{figure*}

The E2C model architecture, figure \ref{fig:method_2D}, consists of a parameterized encoder $\phi_{\theta} : \mathcal{G} \rightarrow \mathcal{Z}$, where $\mathcal{Z} \in \mathbb{R}^{n_z}$, transition layer $\tau_{\theta} : \mathcal{Z} \times \mathcal{U} \rightarrow \mathcal{Z}$ and decoder $\psi_\theta : \mathcal{Z} \rightarrow \mathcal{G}$. The composition of these three gives the overall model $\mathcal{M}_\theta = \{\psi \circ \tau \circ \phi : \mathcal{G} \rightarrow \mathcal{G}\}$. The encoder $\phi$ consists of 4 convolution layers followed by 3 2-layer ResNets and a final dense layer to map from the state space $\mathcal{G}$ to the locally linear latent space $\mathcal{Z}$. The decoder $\psi$ conversely maps back from latent space to state space using a dense layer followed by 3 2-layer transposed ResNets and 4 transposed convolution layers. The transition layer $\tau$ in between aims to model the latent transition function $f^{\text{lat}}$ of the dynamics given by $\textbf{z}_{t+1} = f^{\text{lat}}(\textbf{z}_t, \textbf{u}_t)$. Given the state and action sequences $\textbf{z}_{1:T} = \{\textbf{z}_1,..., \textbf{z}_T\}$ and $\textbf{u}_{1:T} = \{\textbf{u}_1,..., \textbf{u}_T\}$, optimal controls which give rise to the trajectory $\textbf{z}_{1:T}$ for a given $f^{\text{lat}}$ can be determined using traditional optimal control algorithms. These algorithms approximate global non-linear dynamics with local linear dynamics and it can be shown (section 2.2 of Watter, M. et al. \cite{E2C}) that for a reference trajectory $\bar{\textbf{z}}_{1:T}$ and controls $\bar{\textbf{u}}_{1:T}$, the system is linearized as

\begin{center}
    \begin{math}
    \textbf{z}_{t+1} = \textbf{A}(\bar{\textbf{z}}_t)\textbf{z}_t + \textbf{B}(\bar{\textbf{z}}_t)\textbf{u}_t + \textbf{o}(\bar{\textbf{z}}_t),
    \end{math}
\end{center}

where \textbf{A}($\bar{\textbf{z}}_t$) = $\frac{\delta f^{\text{lat}}(\textbf{z}_t, \textbf{u}_t)}{\delta \bar{\textbf{z}}_t}$, \textbf{B}($\bar{\textbf{z}}_t$) = $\frac{\delta f^{\text{lat}}(\textbf{z}_t, \textbf{u}_t)}{\delta \bar{\textbf{u}}_t}$ are local Jacobians and \textbf{o}($\bar{\textbf{z}}_t$) is an offset. This equation is the inspiration for the transition layer and the local Jacobians and offsets are parameterized as trainable weights in the model for calculating $\textbf{z}_{t+1}$.

The loss function of the model is a composition of traditional auto-encoder loss functions and physics-informed loss functions as follows, 
\begin{center}
\begin{math}
    \mathcal{L} = \mathcal{L}_{rec} + \mathcal{L}_{pred} + \theta_{trans} \times \mathcal{L}_{trans} + \theta_{flux} \times \mathcal{L}_{flux} 
\end{math}
\end{center}
\begin{center}
\begin{math} 
    \mathcal{L}_{rec} = \| \textbf{x}_t - \hat{\textbf{x}}_t \|_2, \quad \mathcal{L}_{pred} = \| \textbf{x}_t - \hat{\textbf{x}}_{t+1} \|_2 , \quad \mathcal{L}_{trans} = \| \textbf{z}_{t+1} - \hat{\textbf{z}}_{t+1} \|_2
\end{math}
\end{center}
where $\hat{\textbf{x}}_t = \phi(\psi(\textbf{x}_t))$ is the reconstruction of $\textbf{x}_t$, $\hat{\textbf{x}}_{t+1} = \mathcal{M}(\textbf{x}_t, \textbf{u}_t)$ is the next state prediction, $\hat{\textbf{z}}_{t+1} = \phi(\tau(\textbf{x}_t, \textbf{u}_t))$ is the next state prediction in latent space and $\textbf{z}_{t+1} = \phi(\textbf{x}_{t+1})$ is the latent representation of true state $\textbf{x}_{t+1}$. The physics-informed losses for the model are more specific to the reservoir simulation task, where $\mathcal{L}_{flux}$ aims to minimize the difference in pressure flux passing in and out of a given cell. The flux term is split up into two components, one to  calculate the flux loss for reconstruction and one for prediction as follows,
\begin{center}
\begin{math}
    \mathcal{L}_{flux} = \|\textbf{k}\mathcal{F}^{rec}\|_2 + \|\textbf{k}\mathcal{F}^{pred}\|_2 
\end{math}
\end{center}
\begin{center}
\begin{math}
    \mathcal{F}^{rec} = [k_{ro}(\textbf{S}^w_t)\Delta \textbf{p}_t - k_{ro}(\hat{\textbf{S}}^w_t)\Delta \hat{\textbf{p}}_t] + [k_{rw}(\textbf{S}^w_t)\Delta \textbf{p}_t - k_{rw}(\hat{\textbf{S}}^w_t)\Delta \hat{\textbf{p}}_t] 
\end{math}
\end{center}
\begin{center}
\begin{math}
    \mathcal{F}^{pred} = [k_{ro}(\textbf{S}^w_{t+1})\Delta \textbf{p}_{t+1} - k_{ro}(\hat{\textbf{S}}^w_{t+1})\Delta \hat{\textbf{p}}_{t+1}] + [k_{rw}(\textbf{S}^w_{t+1})\Delta \textbf{p}_{t+1} - k_{rw}(\hat{\textbf{S}}^w_{t+1})\Delta \hat{\textbf{p}}_{t+1}]
\end{math}
\end{center}
where $\textbf{k}$ is the permeability tensor, $k_{ro}$ and $k_{rw}$ are relative permeability fields of the respective components given the saturation field $\textbf{S}_t$ and $\Delta \textbf{p}$ represents pressure drop across adjacent grid cells. Following conservation laws, we require the $\mathcal{L}_{flux}$ to be minimal in order for the model to represent a physical system.

\subsection{E2CO Reduced Order Model}
The E2CO model expands on this by modifying the transition layer to prediction well outputs $\textbf{y}_{t+1}$. It does so by approximating global non-linear dynamics with local linear dynamics similar to the latent representation $\textbf{z}_t$. Here we parametrize two more local Jacobians $\textbf{C}(\bar{\textbf{z}}_t)$ and $\textbf{D}(\bar{\textbf{z}}_t)$ and calculate the well outputs as following

\begin{center}
    \begin{math}
    \hat{\textbf{y}}_{t+1} = \textbf{C}(\bar{\textbf{z}}_t)\hat{\textbf{z}}_{t+1} + \textbf{D}(\bar{\textbf{z}}_t)\textbf{u}_t.
    \end{math}
\end{center}
We thus introduce an additional loss term to train for the well outputs simultaneously given below.
\begin{center}
\begin{math}
    \mathcal{L} = \mathcal{L}_{rec} + \mathcal{L}_{pred} + \theta_{trans} \times \mathcal{L}_{trans} + \theta_{flux} \times \mathcal{L}_{flux} + \theta_{well} \times \mathcal{L}_{well}
\end{math}
\end{center}
\begin{center}
\begin{math}
    \mathcal{L}_{well} = \| \textbf{y}_{t+1} - \hat{\textbf{y}}_{t+1} \|_2,
\end{math}
\end{center}
where $\hat{\textbf{y}}_{t+1}$ is the predicted well output for time $t+1$ and $\textbf{y}_{t+1}$ is the actual well output. Notice here that we are adding additional complexity to the model by trying to predict another quantity so we expect to have a deterioration in accuracy when trying to predict $\textbf{x}_t$. Hence, we use E2CO only when we explicitly require well outputs and employ E2C otherwise for comparing results.

\subsection{Localized Learning for E2C(O)}

\begin{figure*}[!htb]
  \centering
  \includegraphics[scale=0.6, trim=0 2cm 0 0]{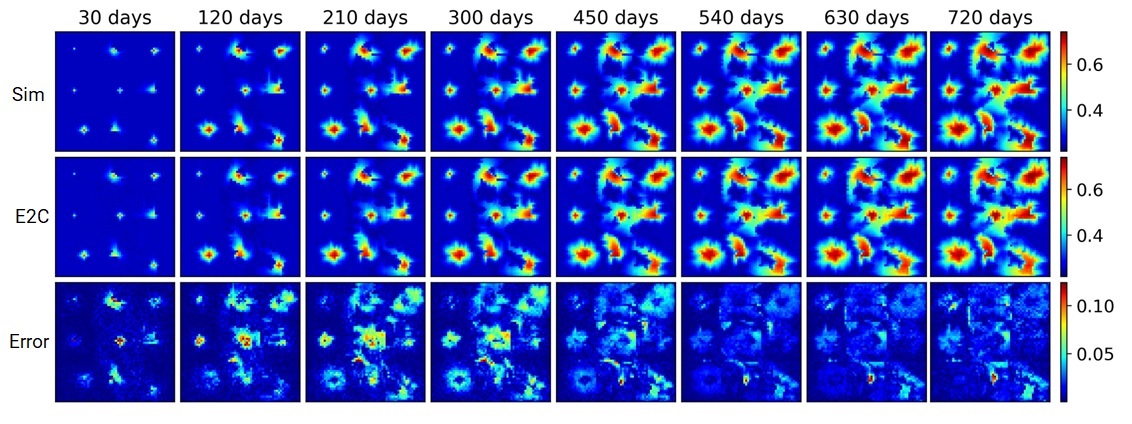}
  \caption{Saturation values predicted by LL-E2C for a single 2D test case. Each time step is 30 days.}
  \label{fig:Results2D}
\end{figure*}


The E2C reduced order model, when scaled to larger grids, was found to grow significantly in training time. For a $60 \times 220 \times 40$ grid, training time exceeds 47.2 hours on an NVIDIA Tesla V100 GPU. To tackle this problem we propose Localized Learning for Embed to Control (LL-E2C) (see figure \ref{fig:method}), a method where, instead of training on a full grid, we train on a random subset of structured sub-grids (sectors).

The initial formulation is as follows. We chose at random input sectors $\textbf{s}_t$ from $\mathcal{S} = \{\textbf{s} : \textbf{s} \in \mathbb{R}^{n_s} \supset \mathcal{G}, \;n_s < n\}$ and train the model to predict sector state $\textbf{s}_{t+1}$. At inference time, we pass the elements of the set of all non-overlapping sectors which are required to fully reconstruct the grid for state $\textbf{s}_t$, $\,\mathcal{I} = \{\textbf{s}^i_t : ||^{n/n_s}_i \textbf{s}^i_t = \textbf{x}_t ,\; \cap^{n/n_s}_i\textbf{s}^i_t = \emptyset\}$, and stitch up the predicted outputs $\textbf{s}^i_{t+1}$ to produce the full next state image $\textbf{x}_{t+1}$.

The key intuition behind sector training is that the underlying physics governing the time evolution of the subsurface properties is constant. This implies that whichever part of the grid we train on, the physics the neural network learns to emulate will remain the same (while respecting boundary conditions), hence sector-wise training loses no generality in solution. 

In reality however, when sectors are chosen for training, we are artificially imposing Dirichlet boundary conditions at the edges where they might otherwise be Von Neumann. For example, when we train using a sector taken from the middle of a grid, all the edges of the sector exhibit Von Neumann boundary conditions but there is no way that the model would be able to capture this without further information. Furthermore, when a sector is chosen from a corner or side, the boundary conditions are a combination of both Dirichlet and Von Neumann. To tackle this problem, we employ a mechanism in which we train on sectors of size $n_s + n_p$ (where $n_p < n_s$), infer on sectors of size $n_s + n_p$ but only use sub-sectors of size $n_s$ to stitch up to the final $\textbf{x}_{t+1}$ state, as shown in figure \ref{fig:bound_method}. The location of the sub-sector is chosen such that its edges line up with the edges of the full image when the inference sector is around the periphery (to capture Dirichlet boundary conditions) and its edges are within the edge of the inference sector when it is taken from the middle of the image (to capture Von Neumann boundary conditions by allowing there to be `flow' out of the sub-sector). We hence modify our formulation to take as inputs expanded sectors from $\Tilde{\mathcal{S}} = \{\Tilde{\textbf{s}} : \Tilde{\textbf{s}} \in \mathbb{R}^{n_s + n_p}\}$, such that $\mathcal{S} \subset {\Tilde{\mathcal{S}}}$ and infer using the set $\,\Tilde{\mathcal{I}} = \{\tilde{\textbf{s}}^i_t : ||^{n/n_s}_i \textbf{s}^i_t = \textbf{x}_t ,\; \cap^{n/n_s}_i\textbf{s}^i_t = \emptyset\}$, where the mapping between $\Tilde{\textbf{s}}^i_t$ and $\textbf{s}^i_t$ is given by the description above.

Another issue which arises from training on sectors is that, although the underlying physics is the same in each sector, the relative locations of the applied controls vary depending on where a sector is extracted from. To tackle this, we add positional encoding by appending the position of the extracted sector and the relative control locations (with the locations of controls outside the sector zeroed out) to the control vector $\textbf{u}_t$. Further to this, permeability information for each cell is passed in as separate channels to provide flow information, aiding the greatly reduced information available to the model.

For the case of predicting well outputs, we extend the above technique to E2CO and propose Localized Learning for Embed to Control and Observe (LL-E2CO), seen in figure \ref{fig:method}. Here we make the same modification to the transition layer as the original E2CO model but now, since we are training on sectors, we need to take into account the fact that each sector only has information about well controls within its boundaries and does not have access to information outside its boundaries. As such, when passing controls $\textbf{u}_t$ into the transition layer to calculate $\textbf{y}_{t+1}$, we zero out the control values of all wells not pertaining to the sector in question. This also acts as an additional proxy for positional encoding of the sector as the model should, in theory, learn the position of the sector based on which well controls are zeroed out.

\section{Results}

\begin{table*}[h]
\caption{Training time results in hours}
\label{tab:results}
\centering
\begin{tabular}{ccccc}
\toprule
\multirow{3}{*}{Method} & \multicolumn{4}{c}{Datasets} \\
\cmidrule{2-5}
& \multicolumn{2}{c}{without Physics-informed Loss} & \multicolumn{2}{c}{with Physics-informed Loss} \\
\cmidrule(lr){2-3}\cmidrule(lr){4-5}
& 2D & 3D & 2D & 3D \\
\midrule
E2C & \textbf{0.4} & 8.3 & 0.7 & 47.2\\
LL-E2C & 0.5 & \textbf{0.9} & \textbf{0.6} & \textbf{2.6}\\
\midrule
E2CO & - & 9.1 & - & 50\\
LL-E2CO & - & 0.9 & - & 2.8\\
\bottomrule
\end{tabular}
\end{table*}

\begin{figure*}[!htb]
  \centering
  \includegraphics[scale=1.25, trim=0 1cm 0 0]{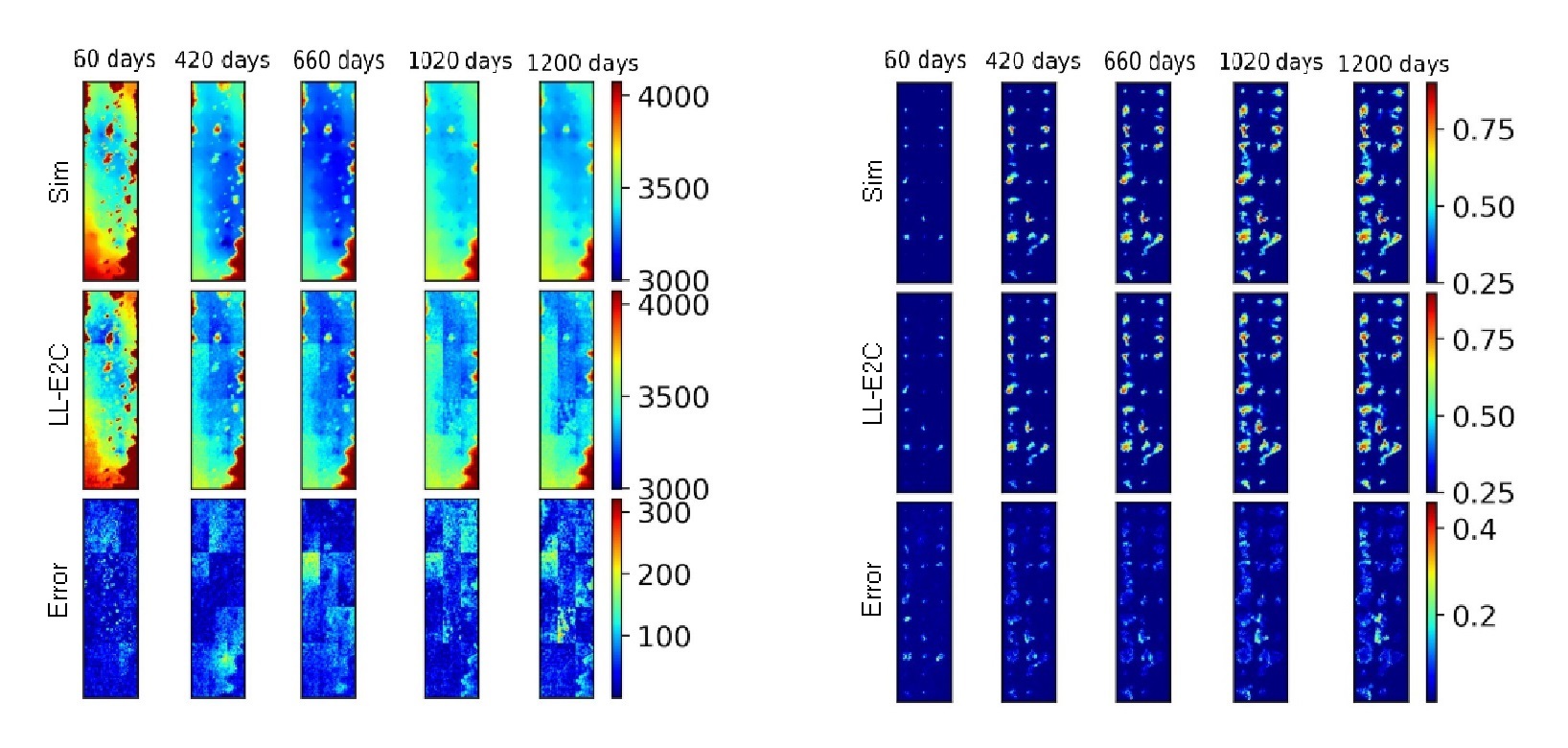}
  \caption{The left plot shows the pressure values from the simulator for a specific layer in the z direction, the predicted pressure values of the LL-E2C model and the absolute error between the two. The right plot shows the same for saturation values. Each time step is 60 days.}
  \label{fig:Results}
\end{figure*}

\begin{figure*}[!htb]
  \centering
  \includegraphics[scale=1.6]{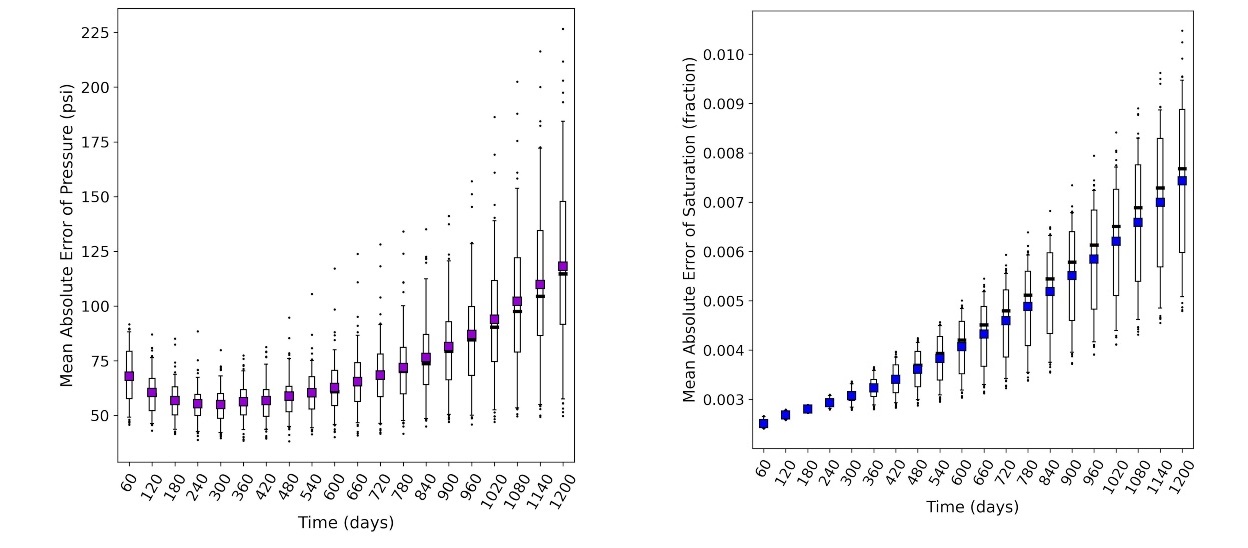}
  \caption{Mean absolute errors of pressure and saturation predictions for the LL-E2C model. Each dot corresponds to an individual test case with the bars signifying 1$\sigma$ confidence intervals. These values correspond to the 3D case with each time step of 60 days.}
  \label{fig:Results_appendix}
\end{figure*}

\begin{figure*}[!htb]
  \centering
  \includegraphics[scale=0.65, trim=0cm 13cm 0cm 0cm]{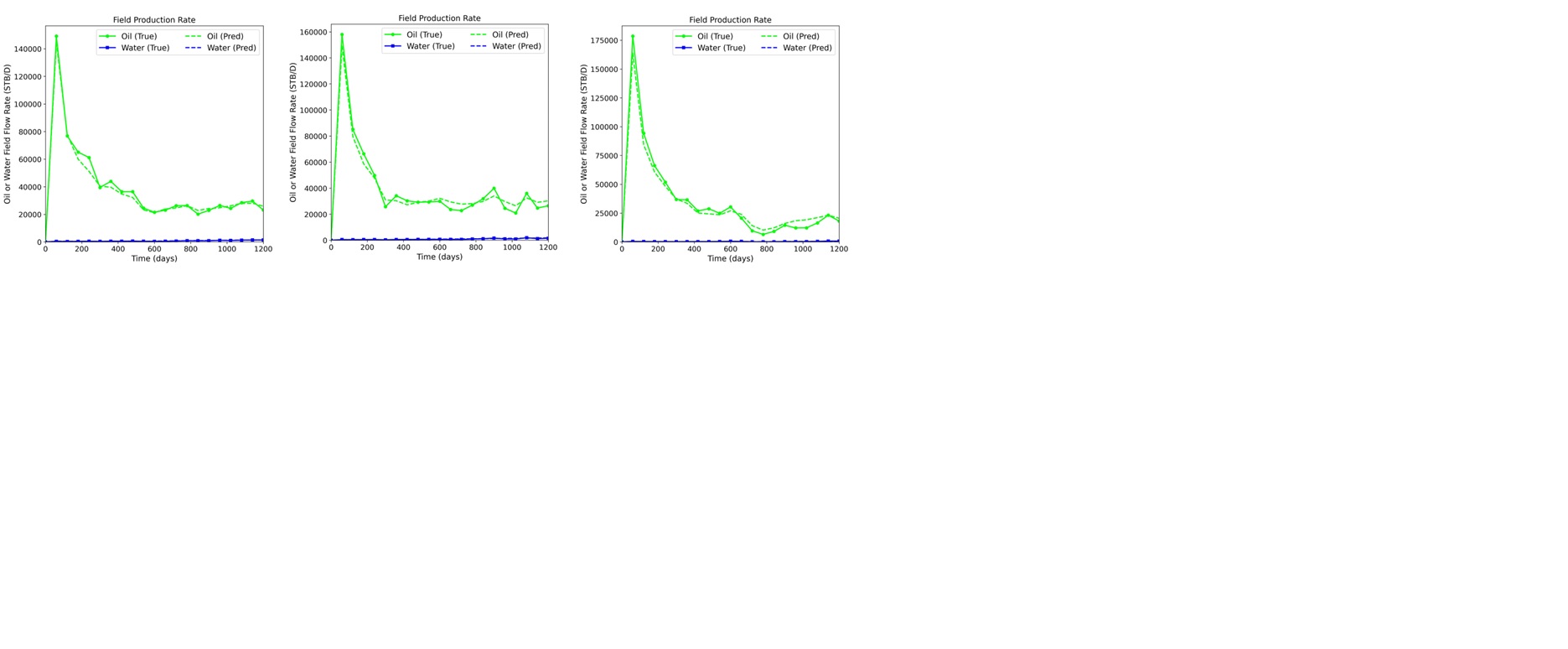}
  \caption{Field production rates of oil and water for 3 different test cases for the LL-E2CO model. The values correspond to the 3D case with each time step corresponding to 60 days.}
  \label{fig:ProdResults}
\end{figure*}

Testing was conducted for the 2D and 3D cases detailed earlier and the respective results are compared to `ground truth' simulation data. In the 2D $60 \times 60$ case, empirically, we found that sector sizes of $(20+4) \times (20+4)$ with an inference size of $20 \times 20$ produced the best results for the LL-E2C model. There was a slight improvement in training times between the E2C and LL-E2C models but we expect to see much a larger difference as we scale up to real world grid sizes. The saturation scalar values for the 2D case are shown in figure \ref{fig:Results2D}. Looking at the errors we see that they mainly peak around the wave fronts of the diffusion process. This can be attributed to the fact that this is where we see the most non-linearity in the system and hence the linearization process taking place in the transition layer does not approximate as well.  

For the 3D $60 \times 220 \times 40$ case, which is much closer to the real world case, with sector sizes of $(20+4) \times (60+8) \times (20+4)$ and an inference size of $20 \times 60 \times 20$, we present a significant speed up time in training from 47.2 hours to 2.6 hours between the E2C and LL-E2C models and 50 hours to 2.8 hours between the E2CO and LL-E2CO models respectively as seen in table \ref{tab:results}. In table \ref{tab:results}, we also see the training times of the models with and without the use of physics-informed losses. Noting that we see a significant increase in accuracy with the inclusion of physics-informed losses, we can conclude that the LL-E2C(O) models are much more powerful for practical use as their training times rival even the E2C(O) models without their physics-informed losses incorporated.

Figure \ref{fig:Results} shows the pressure and saturation predictions for 20 time steps using the LL-E2C model for a slice of the grid in the z direction. By analysing the test errors, we notice that saturation predictions appear to be more accurate than pressure predictions. This is likely due to the fact that pressure is a more global state variable and hence Localized Learning naturally finds it more difficult to model. A more thorough analysis of these errors were carried out and is presented in figure \ref{fig:Results_appendix}. The model was trained and run for all the different test cases and the mean absolute errors in pressure and saturation values are plotted. We observe that the absolute errors in predictions are well within the acceptable range for the purposes of reservoir simulation surrogate modelling, with mean average absolute errors over all test cases ranging between 50 psi to 125 psi for pressure and 0.002 to 0.007 fraction for saturation. We also observe that, although mean average errors increase as we predict further into the future, which is expected due to compounding errors, the model still produces much better results than can be reasonably expected in latter predictions. We attribute this to the fact that the system exhibits maximal non-linearity early on in the time evolution and as the wave fronts of pressure and saturation diffuse out, the transition layer becomes a better approximation of the locally linear dynamics. This leads to a better performance of the LL-E2C model.

When testing the LL-E2CO model, we are mainly concerned with well outputs. As such, in this specific study, we look at the well production rates of oil and water in a reservoir under different well controls. The same analysis can be applied for the $CO_2$ storage task in CCS processes. The model predicts the outputs of each individual well in the system for both the 2D and 3D cases and in figure \ref{fig:ProdResults} we present the field production rates of the 3D case. Field rates are the aggregation of all the well production rates and are a good indication of the accuracy of the model. From the figure we see that the predicted production rates of oil are very close to the actual values produced by the simulator for the 3 cases. At all time steps, and even at later time steps when prediction are expected to diverge more, the values fall within $5\%$ error. This is likely due to the same reasoning mentioned previously for LL-E2C. 

A major advantage of machine learning models over traditional numerical solvers is their ability to parallelize inference by making batchwise computations. We are able to pass multiple test cases simultaneously into our model (which is what we would like to do when trying to find optimal control solutions over a large search space) and extract individual predictions where necessary. Thus, comparing inference times of the proposed LL-E2C(O) with the computation time of HFS, we observe a speed up of the order $10^4$ which brings to light the true strength of machine learning models in this task.


\section{Conclusion}
In this work, we motivate the application of machine learning to subsurface fluid modelling and propose a novel method called Localized Learning to predict future states of an advection-diffusion-reaction system, with only a fraction of the available data. Our method reduces training times by over an order of magnitude for typical 3D cases, while also providing comparable accuracy results for all time steps. In the future, we hope to expand this work to unstructured grids and meshes.

\section{Acknowledgements}
The authors would like to thank SLB management for permission to publish this work. This work was done while Surya T. Sathujoda was a research intern at SLB.

\bibliography{example_paper}

\begin{thebibliography}{21}
\providecommand{\natexlab}[1]{#1}
\providecommand{\url}[1]{\texttt{#1}}
\expandafter\ifx\csname urlstyle\endcsname\relax
  \providecommand{\doi}[1]{doi: #1}\else
  \providecommand{\doi}{doi: \begingroup \urlstyle{rm}\Url}\fi

\bibitem[TPW(2015)]{TPWL5}
\emph{{Retraining Criteria for TPWL/POD Surrogate Based Waterflodding
  Optimization}}, volume Day 3 Wed, February 25, 2015 of \emph{SPE Reservoir
  Simulation Conference}, 02 2015.
\newblock \doi{SPE-173252-MS}.
\newblock D031S011R007.

\bibitem[Atadeger et~al.(2022)Atadeger, Sheth, Vera, Banerjee, and Onur]{E2C3}
Atadeger, A., Sheth, S., Vera, G., Banerjee, R., and Onur, M.
\newblock Deep learning-based proxy models to simulate subsurface flow of
  three-dimensional reservoir systems.
\newblock 2022\penalty0 (1):\penalty0 1--32, 2022.
\newblock ISSN 2214-4609.
\newblock \doi{https://doi.org/10.3997/2214-4609.202244049}.

\bibitem[Bhattacharya et~al.(2021)Bhattacharya, Hosseini, Kovachki, and
  Stuart]{bhattacharya2021model}
Bhattacharya, K., Hosseini, B., Kovachki, N.~B., and Stuart, A.~M.
\newblock Model reduction and neural networks for parametric pdes, 2021.

\bibitem[Cardoso \& Durlofsky(2010)Cardoso and Durlofsky]{TPWL1}
Cardoso, M.~A. and Durlofsky, L.~J.
\newblock Linearized reduced-order models for subsurface flow simulation.
\newblock \emph{J. Comput. Phys.}, 229\penalty0 (3):\penalty0 681–700, 2
  2010.
\newblock ISSN 0021-9991.
\newblock \doi{10.1016/j.jcp.2009.10.004}.

\bibitem[Coutinho et~al.(2021)Coutinho, Dall’Aqua, and Gildin]{E2CO}
Coutinho, E. J.~R., Dall’Aqua, M., and Gildin, E.
\newblock Physics-aware deep-learning-based proxy reservoir simulation model
  equipped with state and well output prediction.
\newblock \emph{Frontiers in Applied Mathematics and Statistics}, 7, 2021.
\newblock ISSN 2297-4687.
\newblock \doi{10.3389/fams.2021.651178}.

\bibitem[He \& Durlofsky(2013)He and Durlofsky]{TPWL3}
He, J. and Durlofsky, L.
\newblock Reduced-order modeling for compositional simulation using trajectory
  piecewise linearization.
\newblock \emph{SPE Journal}, 19, 02 2013.
\newblock \doi{10.2118/163634-MS}.

\bibitem[He \& Durlofsky(2015)He and Durlofsky]{TPWL6}
He, J. and Durlofsky, L.~J.
\newblock Constraint reduction procedures for reduced-order subsurface flow
  models based on pod–tpwl.
\newblock \emph{International Journal for Numerical Methods in Engineering},
  103\penalty0 (1):\penalty0 1--30, 2015.
\newblock \doi{https://doi.org/10.1002/nme.4874}.

\bibitem[He et~al.(2011)He, S\ae{}trom, and Durlofsky]{TPWL2}
He, J., S\ae{}trom, J., and Durlofsky, L.~J.
\newblock Enhanced linearized reduced-order models for subsurface flow
  simulation.
\newblock \emph{J. Comput. Phys.}, 230\penalty0 (23):\penalty0 8313–8341, 9
  2011.
\newblock ISSN 0021-9991.
\newblock \doi{10.1016/j.jcp.2011.06.007}.

\bibitem[He et~al.(2015)He, Zhang, Ren, and Sun]{he2015deep}
He, K., Zhang, X., Ren, S., and Sun, J.
\newblock Deep residual learning for image recognition, 2015.

\bibitem[Jin \& Durlofsky(2018)Jin and Durlofsky]{TPWL7}
Jin, Z.~L. and Durlofsky, L.~J.
\newblock Reduced-order modeling of co2 storage operations.
\newblock \emph{International Journal of Greenhouse Gas Control}, 68:\penalty0
  49--67, 2018.
\newblock ISSN 1750-5836.
\newblock \doi{https://doi.org/10.1016/j.ijggc.2017.08.017}.

\bibitem[Jin et~al.(2020)Jin, Liu, and Durlofsky]{E2C2}
Jin, Z.~L., Liu, Y., and Durlofsky, L.~J.
\newblock Deep-learning-based surrogate model for reservoir simulation with
  time-varying well controls.
\newblock \emph{Journal of Petroleum Science and Engineering}, 192:\penalty0
  107273, 2020.
\newblock ISSN 0920-4105.
\newblock \doi{https://doi.org/10.1016/j.petrol.2020.107273}.

\bibitem[Kovachki et~al.(2023)Kovachki, Li, Liu, Azizzadenesheli, Bhattacharya,
  Stuart, and Anandkumar]{kovachki2023neural}
Kovachki, N., Li, Z., Liu, B., Azizzadenesheli, K., Bhattacharya, K., Stuart,
  A., and Anandkumar, A.
\newblock Neural operator: Learning maps between function spaces, 2023.

\bibitem[Li et~al.(2020{\natexlab{a}})Li, Kovachki, Azizzadenesheli, Liu,
  Bhattacharya, Stuart, and Anandkumar]{li2020multipole}
Li, Z., Kovachki, N., Azizzadenesheli, K., Liu, B., Bhattacharya, K., Stuart,
  A., and Anandkumar, A.
\newblock Multipole graph neural operator for parametric partial differential
  equations, 2020{\natexlab{a}}.

\bibitem[Li et~al.(2020{\natexlab{b}})Li, Kovachki, Azizzadenesheli, Liu,
  Bhattacharya, Stuart, and Anandkumar]{li2020neural}
Li, Z., Kovachki, N., Azizzadenesheli, K., Liu, B., Bhattacharya, K., Stuart,
  A., and Anandkumar, A.
\newblock Neural operator: Graph kernel network for partial differential
  equations, 2020{\natexlab{b}}.

\bibitem[Li et~al.(2021)Li, Kovachki, Azizzadenesheli, Liu, Bhattacharya,
  Stuart, and Anandkumar]{li2021fourier}
Li, Z., Kovachki, N., Azizzadenesheli, K., Liu, B., Bhattacharya, K., Stuart,
  A., and Anandkumar, A.
\newblock Fourier neural operator for parametric partial differential
  equations, 2021.

\bibitem[Li et~al.(2022)Li, Zheng, Kovachki, Jin, Chen, Liu, Azizzadenesheli,
  and Anandkumar]{li2022physicsinformed}
Li, Z., Zheng, H., Kovachki, N., Jin, D., Chen, H., Liu, B., Azizzadenesheli,
  K., and Anandkumar, A.
\newblock Physics-informed neural operator for learning partial differential
  equations, 2022.

\bibitem[Lu et~al.(2021)Lu, Jin, Pang, Zhang, and Karniadakis]{DeepO2021}
Lu, L., Jin, P., Pang, G., Zhang, Z., and Karniadakis, G.~E.
\newblock Learning nonlinear operators via {DeepONet} based on the universal
  approximation theorem of operators.
\newblock \emph{Nature Machine Intelligence}, 3\penalty0 (3):\penalty0
  218--229, mar 2021.
\newblock \doi{10.1038/s42256-021-00302-5}.

\bibitem[Ronneberger et~al.(2015)Ronneberger, Fischer, and
  Brox]{ronneberger2015unet}
Ronneberger, O., Fischer, P., and Brox, T.
\newblock U-net: Convolutional networks for biomedical image segmentation,
  2015.

\bibitem[Rousset et~al.(2014)Rousset, Huang, Klie, and Durlofsky]{TPWL4}
Rousset, M. A.~H., Huang, C.~K., Klie, H., and Durlofsky, L.~J.
\newblock Reduced-order modeling for thermal recovery processes.
\newblock \emph{Computational Geosciences}, 18\penalty0 (3):\penalty0 401--415,
  8 2014.
\newblock ISSN 1573-1499.
\newblock \doi{10.1007/s10596-013-9369-8}.

\bibitem[Watter et~al.(2015)Watter, Springenberg, Boedecker, and
  Riedmiller]{E2C}
Watter, M., Springenberg, J.~T., Boedecker, J., and Riedmiller, M.
\newblock Embed to control: A locally linear latent dynamics model for control
  from raw images.
\newblock In \emph{Proceedings of the 28th International Conference on Neural
  Information Processing Systems - Volume 2}, NIPS'15, pp.\  2746–2754,
  Cambridge, MA, USA, 2015. MIT Press.

\bibitem[Zhu \& Zabaras(2018)Zhu and Zabaras]{ZHU2018415}
Zhu, Y. and Zabaras, N.
\newblock Bayesian deep convolutional encoder–decoder networks for surrogate
  modeling and uncertainty quantification.
\newblock \emph{Journal of Computational Physics}, 366:\penalty0 415--447,
  2018.
\newblock ISSN 0021-9991.
\newblock \doi{https://doi.org/10.1016/j.jcp.2018.04.018}.

\end{thebibliography}
\bibliographystyle{icml2023}

\end{document}